\documentclass[
]{ceurart}

\usepackage{listings}
\usepackage{hyperref}
\usepackage{array}
\usepackage{makecell}
\usepackage{graphicx} 
\begin{document}

\copyrightyear{2025}
\copyrightclause{Copyright for this paper by its authors.
  Use permitted under Creative Commons License Attribution 4.0
  International (CC BY 4.0).}

\conference{CLEF 2025 Working Notes, 9 -- 12 September 2025, Madrid, Spain}

\title{FHSTP@EXIST 2025 Benchmark: Sexism
Detection with Transparent Speech Concept Bottleneck Models}

\title[mode=sub]{Notebook for the EXIST2025 Lab at CLEF 2025}



\author[1]{Roberto Labadie-Tamayo}[%
orcid= 0000-0003-4928-8706,
email=rlabadiet@gmail.com,
]
\fnmark[1]
\cormark[1]

\author[1]{Adrian Jaques Böck}[%
orcid= 0000-0003-1972-0473,
email=Adrian.Boeck@fhstp.ac.at
]
\fnmark[1]
\cormark[1]

\author[1]{Djordje Slijepčević}[%
orcid= 0000-0002-2295-7466,
email=Djordje.Slijepcevic@fhstp.ac.at
]

\author[1]{Xihui Chen}[%
orcid=0000-0002-8131-5092,
email=Xihui.Chen@fhstp.ac.at
]

\author[1]{Andreas Babic}[%
orcid=0009-0008-8927-2593,
email=Andreas.Babic@fhstp.ac.at
]

\author[1]{Matthias Zeppelzauer}[%
orcid= 0000-0003-0413-4746,
email=Matthias.Zeppelzauer@fhstp.ac.at
]

\address[1]{St. P\"{o}lten University of Applied Sciences, 3100 St. Pölten, Austria}

\cortext[1]{Corresponding authors.}
\fntext[1]{These authors contributed equally.}

\begin{abstract}
Sexism has become widespread on social media and in online conversation. 
To help address this issue, the fifth Sexism Identification in Social Networks (EXIST) challenge is initiated at CLEF 2025. Among this year's international benchmarks, we concentrate on solving the first task aiming to identify and classify 
sexism in social media textual posts. 
In this paper, we describe our solutions and report the results 
for all the three subtasks: \textit{Subtask 1.1 -- Sexism Identification in Tweets}, \textit{Subtask 1.2 -- Source Intention in Tweets}, and \textit{Subtask 1.3 -- Sexism Categorization in Tweets}. We implement three models to address each subtask 
which constitute three individual runs, i.e., 
Speech Concept Bottleneck Model (SCBM), Speech Concept Bottleneck Model with Transformer (SCBMT) and a fine-tuned XLM-RoBERTa transformer model. 
SCBM uses descriptive adjectives as human-interpretable
bottleneck concepts. SCBM leverages large language models (LLMs) to encode input texts 
to a human-interpretable representation of descriptive adjectives, which is then used to train a light-weight classifier for downstream tasks. SCBMT extends SCBM by fusing 
the adjective-based representation with the contextual embeddings computed by transformers  to balance interpretability and classification performance.
Beyond the competitive results, these two models offer fine-grained explanations at both the instance level (local) and class level (global).  
We also investigate how additional metadata available in the benchmarks, e.g., annotators' demographic profiles, can be leveraged in our models. 
For the binary classification task (Subtask 1.1), XLM-RoBERTa, fine-tuned on the provided data augmented with datasets from previous years, ranks $6^{\text{th}}$ for English and Spanish content and $4^{\text{th}}$ for English content in the Soft-Soft evaluation. Our SCBMT model achieves $7^{\text{th}}$ place for English and Spanish content and $6^{\text{th}}$ for Spanish content in the Soft-Soft evaluation.




\end{abstract}

\maketitle

\begingroup
\renewcommand\thefootnote{}\footnotetext{Preprint July 2025. 
This version is a preprint uploaded to arXiv. This work has been accepted for presentation 
in the sEXism Identification in Social neTworks (EXIST) task at the 
16\textsuperscript{th} Conference and Labs of the Evaluation Forum (CLEF~2025).}%
\addtocounter{footnote}{-1}
\endgroup

\section{Introduction}
Discriminatory views and statements have increased on social media platforms in recent years, particularly targeting women. This phenomenon, with other issues stemming from social media, e.g., hate speech~\cite{DPS22} and disinformation~\cite{SSSM21}, significantly impacts psychological well-being of those affected and may even lead to physical violence. The EXIST (sEXism Identification in Social neTworks) task at CLEF 2025~\cite{EXIST2025,EXIST2025_WN} aims to advance automatic tools for 
identifying sexism in the multimedia content of social media. We focus on the 
task of sexism identification in textual posts, which forms Task 1 of the EXIST benchmark. 
In this paper, we describe our approaches to the three subtasks, i.e., \textit{Subtask 1.1 -- Sexism Identification in Tweets}, \textit{Subtask 1.2 -- Source Intention in Tweets}, and \textit{Subtask 1.3 -- Sexism Categorization in Tweets} and report the results on the provided data.  

We employ three distinct machine learning (ML) models to address each subtask and provide three different experimental runs for each. 
Our first approach utilizes the \emph{Speech Concept Bottleneck Model} (SCBM), a human-interpretable model that leverages large language models (LLMs) to map input text into an adjective-based representation. 
Originally developed for recognizing online hate and counter speech, 
SCBM offers both global and local explainability. Each dimension of the intermediate representation reflects the degree to which a specific adjective captures an emotion or subjective opinion expressed in the input text. 
A lightweight classifier is then trained on these representations for downstream prediction tasks.
Compared to concept representations based on topics or linguistic patterns, using adjectives as bottleneck concepts significantly improves the interpretability of the model’s outputs.
Our second model, the \emph{Speech Concept Bottleneck Model with Transformer} (SCBMT), extends SCBM by combining the adjective-based representations with transformer-generated embeddings. This fusion enables the model to capture a more comprehensive range of textual features by integrating interpretable and contextual information.
Finally, we fine-tune a multilingual \emph{RoBERTa} transformer model as a baseline to benchmark the performance of our proposed methods.

\smallskip
\noindent\textbf{The structure of the paper.} 
We describe in detail our methodology used to address the challenge,
including the developed models and evaluation settings (e.g., datasets, concept list), in Section~\ref{sec:methodology}. The results are presented in Section~\ref{sec:results}, followed by the discussion and conclusions in Section~\ref{sec:discussion&conclusion}.

\section{Methodology and Evaluation}
\label{sec:methodology}
With the three models employed, we conducted three independent runs for each subtask of the tweet classification challenge of the benchmark. 
In this section, we describe the datasets used, the developed ML models, and the approaches applied for each submitted run of the subtasks. 

\subsection{Datasets}
In addition to the dataset (EXIST2025) of this year's benchmark, 
we utilized data provided in previous years' benchmarks (EXIST2021-EXIST2024). 
The combination increases the scale of training data, and thereby improves model robustness and generalizability. 

\smallskip
\noindent\textbf{EXIST2025 Dataset.} 
The EXIST2025 dataset~\cite{EXIST2025} contains over 10,000 text posts from X/Twitter, written in both English and Spanish, with a balanced distribution between the two languages. The dataset is split into three subsets: training (6,920 posts), development (1,038 posts), and testing (2,076 posts). Each post is annotated by six individuals, and demographic information about the annotators (e.g., gender, age group, ethnicity, education level, and country) is also provided.
Each post is assigned three labels corresponding to the dataset’s three subtasks: \emph{Sexism Identification}, \emph{Source Intention}, and \emph{Sexism Categorization}.
\emph{Subtask 1.1 -- Sexism Identification} is a binary classification task in which each post is labeled as either SEXIST or NON-SEXIST.
\emph{Subtask 1.2 -- Source Intention} is a multiclass classification task that aims to determine the intention behind sexist content, with possible labels being DIRECT, REPORTED, or JUDGEMENTAL.
\emph{Subtask 1.3 -- Sexism Categorization} is a multi-label classification task that identifies specific types of sexist content. The possible categories include: IDEOLOGICAL-INEQUALITY, STEREOTYPING-DOMINANCE, OBJECTIFICATION, SEXUAL-VIOLENCE, and MISOGYNY-NON-SEXUAL-VIOLENCE.

\smallskip\noindent\textbf{EXIST2024 Dataset.} The dataset from the EXIST2024~\cite{EXIST2024} benchmark includes posts from X/Twitter and Gab. We used thetraining set, which consists of 6,920 instances in total, i.e., 3,260 in English and 3,660 in Spanish, and the development set, which includes 1,038 samples. The test set was not used due to missing annotations. 
The class definitions for the tasks of Sexism Identification, Source Intention, and Sexism Categorization are the same as those in the EXIST2025 dataset.

\smallskip\noindent\textbf{EXIST2022 Dataset.} 

We used the training dataset from the EXIST2022 benchmark~\cite{EXIST2022}. This dataset includes posts from X/Twitter and Gab, with a total of 11,345 instances: 5,644 in English and 5,701 in Spanish. The class definitions for the tasks of Sexism Identification and Sexism Categorization are the same as those in the EXIST2025 dataset. The Source Intention task was not included in the benchmark of the EXIST2022 datasets.

\subsection{Lexicon of Descriptive Adjectives for Concept Bottleneck Models}
Considering the specific characteristics of sexist content, we propose a lexicon of descriptive adjectives as the set of concepts for our SCBM and SCBMT models. 
We automatically generate the lexicon by prompting the LLM GPT-o3-mini~\cite{openai2025o3mini}. For each subtask, we provide the LLM with the task definition and example instances, as outlined in~\cite{plaza2025exist}, and ask it to generate 50 adjectives relevant for distinguishing between the classes of the respective classification task.
Thus, the prompts are constructed with in-context learning, consisting of two parts. The first part gives a short description of the classification task, while the second part lists the definitions and examples of the class labels. 
In the following, we demonstrate the prompt used for Task 1.1 as an example 
while the other two prompts can be found in Appendix~\ref{appendix}:

\noindent
\emph{Provide me with 50 adjectives that can be used to describe and distinguish classes in a binary classification task where systems must decide whether or not a given tweet is sexist. The following classes are defined:
\begin{itemize}
    \item Sexist, as in: “Woman driving, be careful!”
    \item  Non-sexist, as in:  “Just saw a woman wearing a mask outside spank her very tightly leashed dog and I gotta say I love learning absolutely everything about a stranger in a single instant.”
\end{itemize}
}
After combining the adjectives from all three subtasks and removing duplicates, we obtained a final set of 132 adjectives (see Table~\ref{tab:adjective-lexicon}).

\begin{table}[ht]
\centering
\caption{Adjective lexicon used as interpretable concepts in SCBM and SCBMT models.}
\label{tab:adjective-lexicon}
\resizebox{\linewidth}{!}{%
  \begin{tabular}{llllll}
\hline
abusive & aggressive & androcentric & antagonistic & antifeminist & anti-egalitarian \\
appearance-driven & assaultive & beauty-obsessed & belittling & belligerent & biased \\
bigoted & body-focused & categorical & censuring & chauvinistic & chronicling \\
coercive & commodifying & condemning & condescending & contemptuous & crass \\
critical & crude & cruel & degrading & dehumanizing & demeaning \\
denigrating & denouncing & depersonalizing & derisive & derogatory & descriptive \\
detailed & devaluing & diminishing & disapproving & discriminatory & disdainful \\
disempowering & dismissive & disparaging & disrespectful & documenting & documenting \\
domineering & eroticizing & essentialist & exclusionary & exploitative & factual \\
generalizing & harassing & hateful & homogenizing & hostile & ignorant \\
incendiary & inegalitarian & inferiorizing & inflammatory & insulting & intolerant \\
invasive & invidious & judgmental & lecherous & lustful & marginalizing \\
masculinist & misandric & misogynistic & mocking & moralizing & narrow-minded \\
non-consensual & objectifying & observational & offensive & oppressive & overgeneralizing \\
oversimplifying & paternalistic & patriarchal & patronizing & pejorative & persecutory \\
predatory & prejudiced & provocative & rapacious & rape-advocating & reactionary \\
rebuking & recounting & reductive & regressive & remonstrative & reporting \\
reproachful & repugnant & repulsive & reviling & rude & scathing \\
scornful & scurrilous & sexist & sexualizing & sexually charged & shallow \\
simplistic & snide & stereotypical & stereotyping & superficial & superiority‑minded \\
threatening & traditionalist & undermining & unjust & venomous & verifiable \\
victim-blaming & victimizing & vilifying & violence-inciting & vitriolic & vituperative \\
\hline
\end{tabular}
}
\end{table}

\subsection{Models}
We implement three models to solve the tasks. The first two models are based on the method we proposed for hate and counter speech recognition.
With its promising performance, we apply two variants of the approach to this new domain of sexism detection. In addition, we employ the multilingual transformer XLM-RoBERTa as a baseline. 

\subsubsection{Speech Concept Bottleneck Model (SCBM)}
SCBM is a deep learning model which we proposed based on the approach of \emph{concept bottleneck models} (CBM) for text classification. Unlike prior models relying on transformers, it provides human-interpretable explanations for the predictions in terms of the most related descriptive adjectives. 
The model is the first to use adjectives as bottleneck concepts to represent input texts, capturing the underlying emotions, intentions, and semantics.  
The SCBM model consists of two sequential steps implemented by two modules: \emph{concept evaluation} and \emph{text classification}. 

\noindent\smallskip \textbf{Concept evaluation.}
This module is used to encode the input text into a representation according to a list of adjectives. For each adjective in Table~\ref{tab:adjective-lexicon}, we prompt an LLM to compute a relevance score between 0 and 1, which evaluates how well the adjective describes the text. Specifically, we query a frozen, 8-bit quantized, pre-trained instance of Llama3.1-8b-instruct~\cite{dubey2024llama, meta_llama3_8b_instruct}  with 
a simple prompt:  \textit{``Tell me if the adjective [adjective]
describes the content of the following text: [text]?''}.
Considering the randomness of LLMs' responses, 
we calculate the relevance score as the marginalized probability of the LLM generating 
a response starting with a positive affirmative word. 
Specifically, we construct a set of words (tokens) that may appear in the prefix of  
an affirmative response, including `Yes', 'Si', and their variants such as '\_yes', `YES'. 
For each word in the set, we obtain the probability for the employed LLM to generate a response starting with that word. 
Then we add up all probabilities of the tokens as the relevance score of the given adjective to the input text. 
%
The output of this module is a vector representation, where each dimension corresponds to the relevance score of an adjective in the list.   

\smallskip\noindent
\textbf{Text classification.}
This module takes the representations of adjective concepts as input and predicts the label(s) of the specific task.
Instead of an one-layer perceptron used in previous CBM-based models, we introduce a relevance gate to further dynamically adjust the importance of each adjective to the final decision. 
The adjusted representation is then given to a multi-layer perceptron (MLP).

To train SCBM, we leverage the EXIST 2025 data and employ a fixed learning rate of 2e-3 with RMSProp, optimized over 300 training epochs. Early stopping is applied based on macro-F1 performance on the validation set.

\subsubsection{Speech Concept Bottleneck Model with Transformer (SCBMT)}
SCBMT is an extended variant of the SCBM model resulting from fusing the adjective-based representation with an embedding of the input text obtained from a transformer model.
The aim is to combine two complementary sources of information, considering that transformer-based models are well known for their capability to capture semantics in the text, while our concept bottleneck representation provides more intention-oriented information about the text.
In our implementation, we use the multilingual transformer, 
XLM-RoBERTa-large~\citep{xlm_roberta_large,Conneau2019UnsupervisedCR}. 
To align the adjective representation with the feature space of the transformer embeddings, we apply a learned linear projection before concatenation.
The SCBMT is trained end-to-end to predict the label(s) of the text defined in the specific task.
%
%
For training SCBMT, we use a fixed learning rate of 1e-5 for all transformer blocks and MLP layers, with optimization performed using RMSProp over 16 epochs.

\subsubsection{Fine-tuned XLM-RoBERTa}
We fine-tune XLM-RoBERTa-large on the EXIST2025 dataset. 
In the fine-tuning process, we additionally incorporate data from previous 
EXIST benchmarks (EXIST2022 and EXIST2024) to increase the amount of training data and thus aim to improve generalizability. 
This fine-tuned XLM-RoBERTa model serves as a  transformer-based baseline. For all three subtasks, we apply the same training parameters, with slight differences in the datasets used. The model is fine-tuned over three epochs with a learning rate of 3e-5, using a linear learning rate scheduler with 500 warm-up steps, weight decay of 0.01, and a batch size of 8. Early stopping is applied based on macro-F1 performance on the validation set.


\subsection{Experiments}
In this section, we present the experimental settings which lead to 
the three submitted experimental runs for each subtask of Task 1. 
Each run explores a distinct ML model. Specifically,  we employ SCBM and SCBMT in Run 1 and  Run 2, respectively, while utilizing the fine-tuned XLM-RoBERTa model in Run 3. 

Due to the specific features of the data in EXIST2025, we have two special considerations when designing each experimental run. 
Recall that six annotators are hired to annotate each tweet, and their demographic information is provided. Our first decision in implementing our experimental run is whether and how to integrate this 
type of information. The second decision is related to the imbalanced data distribution across classes. We  evaluate the impact and select the best strategy to deal with this imbalance. 

\smallskip\noindent \textbf{Demographic information.} 
We explore whether incorporating demographic information can help models better align with the diverse preferences and interpretations of annotators. To investigate this, we evaluated the three ML models under two conditions: i) without incorporating demographic information, and ii) with demographic information included.
In the latter setting, we enriched the prompts to the LLM with each annotator’s demographics as personas, i.e.,  gender, age, ethnicity, educational background, and country. 
In our implementation, one example of this persona information is
`\emph{You are a man aged above 45 years old with latino ethnicity with a Bachelor’s degree coming from Mexico}'.
Since each post is annotated by six individuals, we generated six distinct representations per instance, treating them independently during training. 
At inference time, the model produced six predictions, one for each annotator, and we used majority voting to determine the final prediction.

\smallskip\noindent \textbf{Mitigating class imbalance.} 
Due to class imbalances presented in the dataset (primarily for the multi-class and multi-label classification tasks), we further investigated the 
impact of undersampling techniques as a mitigation strategy. These methods aim to reduce the dominance of majority classes by selectively removing samples, thereby promoting a more balanced class distribution. 

In the following, we outline the specific model architecture used for each run, along with any supplementary methods or configurations applied, such as demographic conditioning, data augmentation (with EXIST datasets from previous years), or data undersampling, which 
contributed to the highest performances on the development set.
We notice that the performance of the deployed models varies across different settings. Moreover, settings with more information incorporated do not always lead to better results. 
According to the exploratory analysis with various settings, we only describe the settings with the best performance in the following.

\subsubsection{Task 1.1: Sexism Identification in Tweets}

\noindent\textbf{Run 1 (SCBM).}
We train the SCBM model without considering annotators' demographic information. 
Additionally, we do not apply undersampling, as the dataset for this subtask is balanced.

\smallskip\noindent \textbf{Run 2 (SCBMT).}
We train the SCBMT model with annotators' demographic information incorporated into the prompts as the persona provided to the LLM. 
For each instance, the model generates six predictions (one per annotator). Majority voting is applied to determine the final output. 
Additionally, no undersampling is need as the data are balanced.

\smallskip\noindent
\textbf{Run 3 (Fine-tuned XLM-RoBERTa).}
We fine-tune an XLM-RoBERTa model on both the training data in EXIST2025  and those from previous years (EXIST2022 and EXIST2024). 
Annotator demographic information does not improve performance, and no undersampling strategies are applied as the data are balanced.

\newpage 
\subsubsection{Task 1.2: Source Intention in Tweets}

\noindent\textbf{Run 1 (SCBM).}
We train the SCBM model without annotators' demographic profiles during training. To address class imbalance, we apply undersampling to the NON-SEXIST class, resulting in a more balanced dataset. 

\smallskip\noindent
\textbf{Run 2 (SCBMT).}
We train the SCBMT model and incorporate annotators' demographic information as the persona in the prompts provided to the LLM. Additionally, we do not apply undersampling.

\smallskip\noindent
\textbf{Run 3 (Fine-tuned XLM-RoBERTa).}
We fine-tune an XLM-RoBERTa model  on the EXIST2025 dataset with undersampling the NON-SEXIST class. No annotators' demographic information is integrated.

\subsubsection{Task 1.3: Sexism Categorization in Tweets}

\noindent \textbf{Run 1 (SCBM).}
We train the SCBM model for multi-label classification without incorporating annotator information or applying any undersampling strategies.

\smallskip\noindent \textbf{Run 2 (SCBMT).}
We train the SCBMT model for multi-label classification without incorporating annotator information or applying any undersampling strategies.

\smallskip\noindent \textbf{Run 3 (Fine-tuned XLM-RoBERTa).}
We fine-tune an XLM-RoBERTa model on the EXIST2025 dataset for multi-label classification, without applying undersampling or incorporating annotator information.

\section{Results}
\subsection{Classification Performances}
\label{sec:results}

\begin{table*}
\centering
\caption{Subtask 1.1 (Sexism Identification in Tweets):  Results for Soft-Soft and Hard-Hard evaluation for English (EN), Spanish (ES) and combinded (ALL) languages.}
\begin{tabular}{cclccccc}
\toprule
Task & Lang & Run & Approach & ICM-Soft & ICM-Soft Norm & Cross Ent. & Rank \\
\midrule
Soft-Soft & ALL & 1 & SCBM & 0.3135 & 0.5503 & 2.0735 & $36^{th}$ \\
Soft-Soft & ALL & 2 & SCBMT & 0.3039 & 0.5487 & 2.524 & $37^{th}$ \\
\textbf{Soft-Soft} & \textbf{ALL} & \textbf{3} & \textbf{XLM-RoBERTa} & \textbf{0.7852} & \textbf{0.6259} & \textbf{0.8416} & \textbf{$6^{th}$} \\
Soft-Soft & ES & 1 & SCBM & 0.542 & 0.5869 & 1.8788 & $26^{th}$ \\
Soft-Soft & ES & 2 & SCBMT & 0.4338 & 0.5696 & 2.4597 & $32^{nd}$ \\
Soft-Soft & ES & 3 & XLM-RoBERTa & 0.7824 & 0.6255 & 0.8276 & $14^{th}$ \\
Soft-Soft & EN & 1 & {SCBM} & -0.0441 & 0.4929 & 2.2921 & $45^{th}$ \\
Soft-Soft & EN & 2 & SCBMT & 0.0679 & 0.5109 & 2.5961 & $42^{nd}$ \\
\textbf{Soft-Soft} & \textbf{EN} & \textbf{3} & \textbf{XLM-RoBERTa} & \textbf{0.7484} & \textbf{0.6202} & \textbf{0.8575} & \textbf{$4^{th}$} \\
\midrule
Hard-Hard & ALL & 1 & SCBM & 0.4288 & 0.7155 & 0.7392 & $105^{th}$ \\
Hard-Hard & ALL & 2 & SCBMT & 0.5545 & 0.7787 & 0.7767 & $21^{st}$ \\
Hard-Hard & ALL & 3 & XLM-RoBERTa & 0.561 & 0.7819 & 0.7839 & $18^{th}$ \\
Hard-Hard & ES & 1 & SCBM & 0.4591 & 0.7296 & 0.7745 & $73^{rd}$ \\
Hard-Hard & ES & 2 & SCBMT & 0.5427 & 0.7714 & 0.7927 & $27^{th}$ \\
Hard-Hard & ES & 3 & XLM-RoBERTa & 0.5573 & 0.7787 & 0.7958 & $20^{th}$ \\
Hard-Hard & EN & 1 & SCBM & 0.3699 & 0.6888 & 0.6887 & $127^{th}$ \\
Hard-Hard & EN & 2 & SCBMT & 0.5512 & 0.7813 & 0.7547 & $26^{th}$ \\
Hard-Hard & EN & 3 & XLM-RoBERTa & 0.5522 & 0.7818 & 0.7678 & $23^{rd}$ \\
\bottomrule
\end{tabular}
\label{tab:task1.1}
\end{table*}

\begin{table*}
\centering
\caption{Subtask 1.2 (Sexism Categorization in Tweets): Results for Soft-Soft and Hard-Hard evaluation for English (EN), Spanish (ES) and combinded (ALL) languages.}
\begin{tabular}{cclccccr}
\toprule
Task & Lang & Run & Approach & ICM-Soft & ICM-Soft Norm & Cross Ent. & Rank \\
\midrule
Soft-Soft & ALL & 1 & SCBM & -1.8291 & 0.3526 & 1.569 & $12^{th}$ \\
\textbf{Soft-Soft} & \textbf{ALL} & \textbf{2} & \textbf{SCBMT} & \textbf{-1.1866} & \textbf{0.4044} & \textbf{1.6566} & \textbf{$7^{th}$} \\
Soft-Soft & ALL & 3 & XLM-RoBERTa & -1.7928 & 0.3556 & 1.7156 & $11^{th}$ \\
Soft-Soft & ES & 1 & SCBM & -1.7971 & 0.3561 & 1.5993 & $16^{th}$ \\
\textbf{Soft-Soft} & \textbf{ES} & \textbf{2} & \textbf{SCBMT} & \textbf{-0.8587} & \textbf{0.4312} & \textbf{1.6174} & \textbf{$6^{th}$} \\
Soft-Soft & ES & 3 & XLM-RoBERTa & -1.5388 & 0.3768 & 1.689 & $12^{th}$ \\
Soft-Soft & EN & 1 & SCBM & -1.9748 & 0.3386 & 1.5351 & $13^{th}$ \\
Soft-Soft & EN & 2 & SCBMT & -1.7874 & 0.3539 & 1.7007 & $10^{th}$ \\
Soft-Soft & EN & 3 & XLM-RoBERTa & -2.3042 & 0.3117 & 1.7455 & $16^{th}$ \\
\midrule
Hard-Hard & ALL & 1 & SCBM & 0.0341 & 0.5111 & 0.4709 & $72^{nd}$ \\
Hard-Hard & ALL & 2 & SCBMT & 0.2795 & 0.5909 & 0.5175 & $16^{th}$ \\
Hard-Hard & ALL & 3 & XLM-RoBERTa & 0.0953 & 0.5310 & 0.4888 & $55^{th}$ \\
Hard-Hard & ES & 1 & SCBM & -0.0351 & 0.4890 & 0.479 & $76^{th}$ \\
Hard-Hard & ES & 2 & SCBMT & 0.3783 & 0.6182 & 0.5443 & $13^{th}$ \\
Hard-Hard & ES & 3 & XLM-RoBERTa & 0.1329 & 0.5415 & 0.5127 & $50^{th}$ \\
Hard-Hard & EN & 1 & SCBM & 0.076 & 0.5263 & 0.4572 & $55^{th}$ \\
Hard-Hard & EN & 2 & SCBMT & 0.1262 & 0.5437 & 0.4769 & $30^{th}$ \\
Hard-Hard & EN & 3 & XLM-RoBERTa & 0.0184 & 0.5064 & 0.453 & $76^{th}$ \\
\bottomrule
\end{tabular}
\label{tab:task1.2}
\end{table*}

\begin{table*}
\centering
\caption{Subtask 1.3 (Sexism Categorization in Tweets ): Results for Soft-Soft and Hard-Hard evaluation for English (EN), Spanish (ES) and combinded (ALL) languages. Runs with missing results are noted with "-".}
\begin{tabular}{cclccccr}
\toprule
Task & Lang & Run & Approach & ICM-Soft & ICM-Soft Norm & Cross Ent. & Rank \\
\midrule
Soft-Soft & ALL & 1 & SCBM & -7.9676 & 0.0793 & -- & $24^{th}$ \\
\textbf{Soft-Soft} & \textbf{ALL} & \textbf{2} & \textbf{SCBMT} & \textbf{-3.673} & \textbf{0.3060} & \textbf{--} & \textbf{$14^{th}$} \\
Soft-Soft & ALL & 3 & XLM-RoBERTa & -11.2121 & 0.0000 & -- & $31^{st}$ \\
Soft-Soft & ES & 1 & SCBM & -8.0491 & 0.0811 & -- & $23^{rd}$ \\
\textbf{Soft-Soft} & \textbf{ES} & \textbf{2} & \textbf{SCBMT} & \textbf{-3.3946} & \textbf{0.3233} & \textbf{--} & \textbf{$13^{th}$} \\
Soft-Soft & ES & 3 & XLM-RoBERTa & -10.7122 & 0.0000 & -- & $30^{th}$ \\
Soft-Soft & EN & 1 & SCBM & -8.0029 & 0.0615 & -- & $25^{th}$ \\
Soft-Soft & EN & 2 & SCBMT & -4.0944 & 0.2757 & -- & $19^{th}$ \\
Soft-Soft & EN & 3 & XLM-RoBERTa & -11.9457 & 0.0000 & -- & $31^{st}$ \\
\midrule
Hard-Hard & ALL & 1 & SCBM & -0.0825 & 0.4809 & 0.5461 & $43^{rd}$ \\
Hard-Hard & ALL & 2 & SCBMT & 0.1852 & 0.5430 & 0.5700 & $19^{th}$ \\
Hard-Hard & ALL & 3 & XLM-RoBERTa & 0.0978 & 0.5227 & 0.5728 & $28^{th}$ \\
Hard-Hard & ES & 1 & SCBM & -0.1690 & 0.4623 & 0.5485 & $49^{th}$ \\
Hard-Hard & ES & 2 & SCBMT & 0.2467 & 0.5551 & 0.5859 & $19^{th}$ \\
Hard-Hard & ES & 3 & XLM-RoBERTa & 0.1566 & 0.5350 & 0.5878 & $25^{th}$ \\
Hard-Hard & EN & 1 & SCBM & -0.0144 & 0.4965 & 0.5380 & $46^{th}$ \\
Hard-Hard & EN & 2 & SCBMT & 0.1044 & 0.5256 & 0.5498 & $29^{th}$ \\
Hard-Hard & EN & 3 & XLM-RoBERTa & 0.0140 & 0.5034 & 0.5517 & $41^{st}$ \\
\bottomrule
\end{tabular}
\label{tab:task1.3}
\end{table*}

The results across all three subtasks show clear differences in performance depending on the task and language setting. In Subtask 1.1 (Table~\ref{tab:task1.1}), XLM-RoBERTa performed best overall, particularly in the Soft-Soft evaluation, achieving top ranks and the lowest Cross Entropy scores across languages. Our best submission was in this subtask, where the XLM-RoBERTa model achieved the $4^{th}$ place in the overall benchmark. In this subtask, SCBM and SCBMT performed almost the same in the Soft-Soft evaluation, with SCBMT slightly outperforming SCBM. However, in the Hard-Hard evaluation, SCBM performed significantly worse compared to SCBMT by about 2-6\% in Cross Entropy and 4-10\% in ICM-Soft Norm across both languages.

For Subtask 1.2 (Table~\ref{tab:task1.2}), SCBMT showed stronger results compared to Subtask 1.1, where it achieved the largest ICM-Soft scores in the Soft-Soft evaluation,  ranking the $6^{th}$ place in the benchmark for English. It also performed comparably well on the mixed English-Spanish dataset, securing the $7^{th}$ place. XLM-RoBERTa was less consistent than in the previous task, particularly in the Hard-Hard evaluation. The SCBM performed comparably to XLM-RoBERTa but overall considerably worse than the combined approach SCBMT.

Subtask 1.3 (Table~\ref{tab:task1.3}) was the most challenging task across all models. SCBMT again yielded the most reliable results, ranking $13^{\text{th}}$ for Spanish in the Soft-Soft evaluation. SCBM performed better in the Soft-Soft evaluation compared to XLM-RoBERTa but performed slightly worse in the Hard-Hard evaluation.

Overall, XLM-RoBERTa performed best in the binary classification setting of Subtask 1.1, especially when fine-tuned with additional social media datasets from previous years. SCBMT performed better for finer-grained categorization of Subtask 1.2 and 1.3 than the XLM-RoBERTa model (especially for Spanish data). SCBM demonstrates the lowest performance in most cases, although its performance in the Soft-Soft evaluation was comparable to that of XLM-RoBERTa (except for in Subtask 1.1).  

\subsection{Interpretability of SCBM}

\begin{table}[t]
    \centering
    \caption{Local explanation for randomly selected correctly classified instances.}
    \label{tab:local-exp}
\resizebox{\linewidth}{!}{%
\begin{tabular}{>{\centering\arraybackslash}m{1.2cm} >{\centering\arraybackslash}m{1.2cm} >{\raggedright\arraybackslash}m{4.5cm} >{\raggedright\arraybackslash}m{6cm} >{\raggedright\arraybackslash}m{6.5cm}}
\hline
\textbf{Lang} & \textbf{Task} & \textbf{Class} & \textbf{Tweet} & \textbf{Explanation} \\
\hline
\multirow{1}{*}{EN} & \multirow{1}{*}{Task 1.1}& SEXIST & @ElDukemane You look like a whore in ur new picture. Love to see it. & victim‑blaming, harassing, devaluing, insulting, oppressive, sexist, misogynistic, appearance‑driven, predatory, hostile \\
\midrule
\multirow{8}{*}{EN} & \multirow{8}{*}{Task 1.2} & DIRECT & @LaceyRoseeeeeee @OliveCelesteof Eat out both holes and then slide my dick in you and pound you till you squirt & androcentric, invasive, patriarchal, violence‑inciting, dehumanizing, abusive, rape‑advocating, assaultive, hostile, predatory \\
 & & JUDGEMENTAL & Amazing that the GOP is trying to take away our ability to dispose of a bundle of cells... & patriarchal, provocative, chauvinistic, inflammatory, masculinist, simplistic, homogenizing, repugnant, victim‑blaming, traditionalist \\
 & & REPORTED & BOOK REVIEW: “Sexism dictates the lives of most women... \#KimJiyoungBorn1982 \#sexualassault” & critical, incendiary, disempowering, moralizing, inflammatory, biased, denouncing, reductive, provocative, depersonalizing \\
\midrule
\multirow{14}{*}{EN} & \multirow{14}{*}{Task 1.3}& IDEOLOGICAL - INEQUALITY & @johnrclark12 Yet, there's a DIRECT correlation between the “women's revolution” and declining ratings... & simplistic, paternalistic, essentialist, stereotypical, traditionalist, narrow‑minded, intolerant, dehumanizing, exclusionary, non‑consensual \\
 & & MISOGYNY NON‑SEXUAL VIOLENCE & why did they make el feel bad for hitting that bitch with a rollerskate? she deserved that shit & vitriolic, intolerant, devaluing, denigrating, invasive, belittling, venomous, body‑focused, offensive, inflammatory \\
 & & OBJECTIFICATION & “Get changed, you look like a prostitute.” She smirked and walked off... & superficial, body‑focused, critical, devaluing, rude, scornful, oppressive, disrespectful, depersonalizing, reactionary \\
 & & SEXUAL‑VIOLENCE & @Fox\_x\_Gods Maybe something involving her getting gangbanged? & rude, invasive, crass, dehumanizing, belittling, crude, marginalizing, scurrilous, repugnant, pejorative \\
 & & STEREOTYPING‑DOMINANCE & @DAZNBoxing Something deep inside of me finds it difficult to watch women’s boxing... & critical, traditionalist, androcentric, anti‑egalitarian, categorical, generalizing, biased, patriarchal, depersonalizing, stereotypical \\
\midrule
\multirow{1}{*}{ES} & \multirow{1}{*}{Task 1.1} & SEXIST & @carolinagomezsn Así debe ser, una mujer obediente... & misogynistic, predatory, anti‑egalitarian, antagonistic, rape‑advocating, appearance‑driven, assaultive, sexualizing, disdainful, exploitative \\
\midrule
\multirow{8}{*}{ES} &  \multirow{8}{*}{Task 1.2} & DIRECT & @FlorenciaLagosN Andate a Cuba, yo te pago los pasajes… & chauvinistic, discriminatory, patriarchal, misogynistic, dismissive, dehumanizing, intolerant, androcentric, bigoted, inferiorizing \\
 & & JUDGEMENTAL & Esta publicidad solo incluye a gais y bisexuales hombres cisgénero... & victim‑blaming, inferiorizing, victimizing, reproachful, belligerent, dismissive, generalizing, patriarchal, homogenizing, derogatory \\
 & & REPORTED & En nuestra provincia, existe una importante brecha de género... & generalizing, victim‑blaming, victimizing, homogenizing, inflammatory, patriarchal, essentialist, dismissive, inferiorizing, prejudiced \\
\midrule
\multirow{14}{*}{ES} & \multirow{14}{*}{Task 1.3} & IDEOLOGICAL - INEQUALITY & @WillyTolerdoo @MeerRocio Yo es que soy “masculinista”... & overgeneralizing, moralizing, reductive, patronizing, homogenizing, biased, censuring, categorical, condescending, dehumanizing \\
 & & MISOGYNY NON‑SEXUAL VIOLENCE & @Mzavalagc tienes cara de mojigata, pero detrás hay una mujer muy corrupta... & belligerent, body‑focused, intolerant, essentialist, critical, paternalistic, devaluing, misogynistic, invasive, bigoted \\
 & & OBJECTIFICATION & Un hombre con plata es como una mujer bonita... & commodifying, bigoted, domineering, paternalistic, dehumanizing, diminishing, victimizing, exploitative, patriarchal, beauty‑obsessed \\
 & & SEXUAL‑VIOLENCE & @girldomf me gustaría follar a alguna de ustedes... & invasive, rude, belligerent, pejorative, offensive, body‑focused, crass, denigrating, disdainful, rape‑advocating \\
 & & STEREOTYPING‑DOMINANCE & @Maria\_RuizG1 Ojalá no vote, porque las mujeres no deberían votar :)... & victimizing, belligerent, intolerant, bigoted, ignorant, assaultive, disempowering, rape‑advocating, paternalistic, essentialist \\
 \hline
\end{tabular}%
}
\end{table}

We observe that SCBM when applied alone is less effective than the other two models which benefit from text representations calculated from 
encoder-based transformers. However, the use of adjective concepts makes the model inherently interpretable. 
As previously mentioned, SCBM implements a trainable relevance gate to further adjust the importance of adjective concepts and generate the final input encoding for classification. This representation serves as the explanation of SCBM.
In this section, we will examine the quality of these adjective-based explanations from both the global and local level.

To assess the local interpretability of SCBM, we manually examine the top 10 adjectives with the largest relevance scores.
In \tableautorefname~\ref{tab:local-exp}, we randomly select one correctly classified tweet from each class defined in every subtask, and present the top 10 relevant adjectives.  
Across all three tasks, we observe that in general, the SCBM produces semantically rich, domain-relevant adjective explanations. The majority of these adjectives support the classification outcome. Furthermore, they also expose underlying discursive features of sexist content. 

For instance, in Subtask 1.2, the model consistently differentiates stylistic and intentional nuances. For example, tweets labeled as DIRECT are associated with adjectives such as \textit{invasive}, \textit{androcentric}, and \textit{violence-inciting}, revealing overt and aggressive narrative strategies. In contrast, the JUDGEMENTAL class leans on subtler descriptors like \textit{chauvinistic}, \textit{provocative}, and \textit{victim-blaming}, pointing toward implicit bias. Meanwhile, the REPORTED category is likely to be described by adjectives like \textit{critical}, \textit{ incendiary}, and \textit{moralizing}, indicating second-hand commentary on sexist events, which aligns with the meta-discursive nature of such content.


\begin{table}[t]
    \centering
    \caption{Global explanations for multi-label and multiclass classification tasks.}
    \label{tab:global-exp}
    \fontsize{8pt}{9pt}\selectfont
\resizebox{\linewidth}{!}{%
\begin{tabular}{>{\centering\arraybackslash}m{1.2cm} >{\centering\arraybackslash}m{2cm} >{\raggedright\arraybackslash}m{3.5cm} >{\raggedright\arraybackslash}m{8cm}}
\hline
\textbf{Lang} & \textbf{Task} & \textbf{Class} & \textbf{Explanation} \\
\hline
\multirow{6}{*}{EN} & \multirow{6}{*}{Task 1.2} & DIRECT & patriarchal, dehumanizing, chauvinistic, androcentric, vitriolic, inflammatory, bigoted, anti‑egalitarian, derogatory, homogenizing \\
 &  & JUDGEMENTAL & victim-blaming, inflammatory, homogenizing, patriarchal, chauvinistic, antagonistic, inegalitarian, dismissive, generalizing, victimizing \\
 &  & REPORTED & inflammatory, critical, homogenizing, patriarchal, documenting, generalizing, disempowering, inegalitarian, incendiary, inferiorizing \\
 \midrule
 \multirow{9}{*}{EN} & \multirow{9}{*}{Task 1.3} & IDEOLOGICAL-INEQUALITY & critical, androcentric, belligerent, paternalistic, provocative, traditionalist, body-focused, ignorant, victimizing, intolerant \\
 &  & MISOGYNY-NON-SEXUAL-VIOLENCE & belligerent, intolerant, body-focused, antifeminist, critical, simplistic, assaultive, paternalistic, invasive, ignorant \\
 &  & OBJECTIFICATION & invasive, belligerent, intolerant, paternalistic, antifeminist, oppressive, domineering, body-focused, bigoted, essentialist \\
 &  & SEXUAL-VIOLENCE & body-focused, invasive, belligerent, intolerant, antifeminist, pejorative, superficial, simplistic, misogynistic, critical \\
 &  & STEREOTYPING-DOMINANCE & paternalistic, belligerent, critical, intolerant, domineering, victimizing, traditionalist, commodifying, androcentric, bigoted, stereotypical \\
 \midrule
\multirow{6}{*}{ES} & \multirow{6}{*}{Task 1.2} & DIRECT & patriarchal, dehumanizing, chauvinistic, androcentric, vitriolic, inflammatory, bigoted, anti‑egalitarian, derogatory, homogenizing \\
 &  & JUDGEMENTAL & victim-blaming, inflammatory, homogenizing, patriarchal, chauvinistic, antagonistic, inegalitarian, dismissive, generalizing, victimizing \\
 &  & REPORTED & inflammatory, critical, homogenizing, patriarchal, documenting, generalizing, disempowering, inegalitarian, incendiary, inferiorizing \\
 \midrule
 \multirow{9}{*}{ES} & \multirow{9}{*}{Task 1.3} & IDEOLOGICAL-INEQUALITY & critical, androcentric, belligerent, paternalistic, provocative, traditionalist, body-focused, ignorant, victimizing, intolerant \\
 &  & MISOGYNY-NON-SEXUAL-VIOLENCE & belligerent, intolerant, body-focused, antifeminist, critical, simplistic, assaultive, paternalistic, invasive, ignorant \\
 &  & OBJECTIFICATION & invasive, belligerent, intolerant, paternalistic, antifeminist, oppressive, domineering, body-focused, bigoted, essentialist \\
 &  & SEXUAL-VIOLENCE & body-focused, invasive, belligerent, intolerant, antifeminist, pejorative, superficial, simplistic, misogynistic, critical \\
 &  & STEREOTYPING-DOMINANCE & paternalistic, belligerent, critical, intolerant, domineering, victimizing, traditionalist, commodifying, androcentric, bigoted \\
 \hline
\end{tabular}
}
\end{table}

To provide a global explanation for SCBM, we aggregate local explanations to identify the adjectives that contribute the most to predicting specific classes. The global contribution of an adjective to a particular class is computed as the mean value of its activations across all correctly classified instances in the training set belonging to that class. \tableautorefname~\ref{tab:global-exp} shows the global explanation obtained for the multi-label and multi-class classification tasks.
The global explanations provided by SCBM align strongly with the patterns observed in local explanations, reinforcing the model’s consistency in how it maps adjective concepts to classes. Frequently occurring adjectives like \textit{paternalistic}, \textit{intolerant}, and \textit{body-focused} appear both at the instance level and across entire classes, confirming that the model’s decisions are grounded in stable, semantically relevant features.

However, we also observe meaningful overlap between classes, particularly within ideologically adjacent labels. For example, adjectives such as \textit{belligerent} and \textit{invasive} appear in both SEXUAL-VIOLENCE and OBJECTIFICATION, while \textit{critical} and \textit{androcentric} are shared across IDEOLOGICAL-INEQUALITY and STEREOTYPING-DOMINANCE. This reflects the fuzzy boundaries between sexism subtypes and highlights the complexity of capturing such discourse through discrete labels.

\newpage
\section{Discussion \& Conclusion}
\label{sec:discussion&conclusion}


XLM-RoBERTa performs best in the binary classification task (Subtask 1.1), especially when fine-tuned with additional topic-related data. Our SCBMT model, however, proves more suitable for fine-grained multilabel classification problems by outperforming fine-tuned transformers. This outcome can be directly attributed to the complementary semantic information captured by our adjective-based representation. Another advantage of our SCBMT model over the plain transformer approach is that the transformer component in SCBMT is not fine-tuned, offering a lightweight alternative to the resource-intensive fine-tuning typically required by transformer models. In addition, both the SCBM and the SCBMT model provide high interpretability. 

Interestingly, although our adjectives and prompts were only designed for English, SCBMT still performed comparably well on Spanish data. In some cases, SCBMT even achieved better performance on the Spanish data. This suggests a high degree of language robustness of our method. This robustness may be attributed not only to the multilingual capabilities of large language models, but also to the generalizability of the semantic cues captured by our adjective-based concepts.
A promising direction for future work involves employing more sophisticated prompting techniques to derive the adjective-based representation. Furthermore, collaboration with domain experts could support the expansion of the automatically generated adjective lexicon in order to further improve its coverage and interpretability for the intended users. 

\begin{acknowledgments}
  This work has been funded by the Vienna Science and Technology Fund (WWTF) [10.47379/ICT20016] and by the Austrian Research Promotion Agency FFG under project no. 898085 (project TrustAI) and FO999904624 (project FairAI).
\end{acknowledgments}

\section*{Declaration on Generative AI}
 During the preparation of this work, the authors used ChatGPT in order to: Grammar and spelling check. After using this service, the authors reviewed and edited the content as needed and take full responsibility for the publication’s content. 

\bibliography{sample-ceur}

\appendix

\section{Prompts used to generate adjective lexicon for Task 1.2 \& 1.3}
\label{appendix}
\subsection{Task 1.2}
\noindent
\emph{
Provide me with 50 adjectives that can be used to describe and distinguish classes in a task that aims to categorize the message according to the intention of the author. It is a ternary classification task: (i) direct sexist message, (ii) reported sexist message and (iii) judgmental message. This distinction allows to differentiate sexism that is actually taking place in online platforms from sexism which is being suffered by women in other situations but that is being reported in social networks with the aim of complaining and fighting against sexism. The following classes are defined:
\begin{itemize}
\item  Direct sexist message: the intention was to write a message that is sexist by itself or incites to be sexist, as in: ``A woman needs love, to fill the fridge, if a man can give this to her in return for her services (housework, cooking, etc), I don’t see what else she needs."
\item  Reported sexist message: the intention is to report and share a sexist situation suffered by a woman or women in first or third person, as in: ``Today, one of my year 1 class pupils could not believe he’d lost a race against a girl."
\item Judgmental message: the intention is judgmental, since the tweet describes sexist situations or behaviors with the aim of condemning them, as in: ``21st century and we are still earning 25\% less than men \#Idonotrenounce."
\end{itemize}
} 
\subsection{Task 1.3}
\noindent
\emph{
Many facets of a woman’s life may be the focus of sexist attitudes including domestic and parenting roles, career opportunities, sexual image, and life expectations, to name a few. Provide me with 50 adjectives that can be used to describe and distinguish following classes:
\begin{itemize}
\item   Ideological and inequality: this category includes tweets that discredit the feminist movement in order to devalue and defame the struggle of women in any aspect of their lives. It also includes tweets that reject inequality between men and women, or present men as victims of gender-based oppression. As in: ``Think the whole equality thing is getting out of hand. We are different, that’s how we’re made!"
\item  Stereotyping and dominance: this category includes messages that express false ideas about women that suggest they are more suitable or inappropriate for certain tasks. It also includes any claim that implies that men are somehow superior to women. As in: ``Most women no longer have the desire or the knowledge to develop a high quality character, even if they wanted to."
\item  Objectification: Objectification and physical stereotyping include messages where women are presented as objects apart from their dignity and personal aspects. We also include messages that assume or describe certain physical qualities that women must have in order to fulfill traditional gender roles, for example, ideas that suggest that women should maintain a standard and ideal of beauty or attacks on a woman’s body. As in: ``No offense but I’ve never seen an attractive African American hooker. Not a single one."
\item  Sexual violence: this category includes messages where sexual suggestions, requests or harassment of a sexual nature (rape or sexual assault) are made, as in: ``I wanna touch your tits..you can’t imagine what I can do on your body."
\item Misogyny and non-sexual violence: this category includes expressions of hatred and violence towards women, as in: ``Domestic abuse is never okay... Unless your wife is a bitch."
\end{itemize}
}
\end{document}